\documentclass{ifacconf}

\usepackage{graphicx}      
\usepackage{natbib}        

\usepackage{bm}
\usepackage{amsmath}
\usepackage{amsfonts, verbatim}
\usepackage{amssymb}            
\usepackage{bbm}                
\usepackage{booktabs}           
\usepackage{wrapfig}
\usepackage{float}
\usepackage{algorithm}
\usepackage[noend]{algpseudocode}
\usepackage[normalem]{ulem}
\DeclareMathAlphabet{\mathpzc}{OT1}{pzc}{m}{it}
\usepackage{xcolor}
\usepackage[normalem]{ulem} 
\usepackage{mathtools}
\usepackage{color}
\usepackage{graphicx}
\usepackage{setspace}

\usepackage{tabularx,ragged2e,booktabs}
\usepackage{enumerate}
\usepackage{multirow}
\usepackage{longtable}
\usepackage{mathtools}

\newcommand{\mbf}[1]{\boldsymbol{#1}}

\newcommand{\inp}[1]{\langle{#1}\rangle}

\newcommand{\abs}[1]{\big| #1 \big|}

\newcommand{\norm}[1]{\left\| #1 \right\|}
\newcommand{\R}{\mathbb{R}}

\newcommand{\bu}{\mbf{u}}
\newcommand{\bv}{\mbf{v}}

\newcommand{\bx}{\mbf{x}}

\newcommand{\by}{\mbf{y}}

\newcommand{\bz}{\mbf{z}}
\newcommand{\bvar}{\mbf{\xi}}

\newcommand{\mE}{\mathcal{E}}

\newcommand{\hypspace}{\mathcal{H}}

\newcommand{\E}{\mathbb{E}}

\newcommand{\argmin}[1]{\underset{#1}{\operatorname{arg}\operatorname{min}}\;}

\begin{document}
\begin{frontmatter}

\title{Learning Interaction Variables and Kernels from Observations of Agent-Based Systems \thanksref{footnoteinfo}} 

\thanks[footnoteinfo]{MM was partially supported by AFOSR FA9550-20-1-0288 and FA9550-21-1-0317, DE‐SC0021361, NSF IIS-1837991 and DMS-1913243.}

\author[a]{Jinchao Feng}
\author[a,b,c]{Mauro Maggioni}
\author[c]{Patrick Martin}
\author[d]{Ming Zhong}

\address[a]{Department of Applied Mathematics $\&$ Statistics, Johns Hopkins University, Baltimore, MD $21218$, USA (e-mail: jfeng34@jhu.edu).}
\address[b]{(e-mail: mauromaggionijhu@icloud.com)}
\address[c]{Department of Mathematics, Johns Hopkins University, Baltimore, MD $21218$, USA (email: mmart152@alumni.jh.edu)}
\address[d]{Texas A$\&$M Institute of Data Science, Texas A$\&$M University, College Station, TX $77843$, USA (e-mail: mingzhong@tamu.edu)}

\begin{abstract}
Dynamical systems across many disciplines are modeled as interacting particles or agents, with interaction rules that depend on a very small number of variables (e.g. pairwise distances, pairwise differences of phases, etc...), functions of the state of pairs of agents. Yet, these interaction rules can generate self-organized dynamics, with complex emergent behaviors (clustering, flocking, swarming, etc.).  We propose a learning technique that, given observations of states and velocities along trajectories of the agents, yields both the variables upon which the interaction kernel depends and the interaction kernel itself, in a nonparametric fashion. This yields an effective dimension reduction which avoids the curse of dimensionality from the high-dimensional observation data (states and velocities of all the agents).  We demonstrate the learning capability of our method to a variety of first-order interacting systems.
\end{abstract}


\end{frontmatter}

\section{Introduction}
Systems of interacting particles or agents are ubiquitous in science.  They may be employed to model and interpret intriguing natural phenomena, such as superconductivity, $n$-body problems, and self-organization in biological systems.  Collective dynamical systems demonstrate emergent behaviors such as clustering (\cite{Krause2000, MT2014}), flocking (\cite{Vicsek_model, CS2007}), milling (\cite{Carrillo2009, Albi2014}), synchronization (\cite{Strogatz2000, OKeeffe2017}), and many more (see \cite{Degond2013}).  

We consider interacting-agent systems of $N$ agents governed by the equations in the form
\begin{equation}
\dot\bx_i(t) = \frac{1}{N}\sum_{i' = 1}^N\Phi(\bx_i(t), \bx_{i'}(t))(\bx_{i'}(t) - \bx_i(t))\,,
\label{eq:first_order}
\end{equation}
for $i = 1, \ldots, N$, $t \in [0, T]$; $\bx_i \in \R^d$ is a state vector, for example describing position, speed, emotion, or opinions of the $i^{th}$ agent, and $\Phi:\R^d\times\R^d \rightarrow \R$ is the \textbf{interaction kernel}, governing how the state of agent $i'$ influences the state of agent $i$.  

Note that while the state space of the system is $dN$-dimensional, the interaction kernel is a function of $2d$ dimensions.
In fact, in many important models, $\Phi$ depends on an even smaller number of natural variables $\bvar\in\mathbb{R}^{d'}$, with $d'\ll 2d$, which are functions of pairs $(\bx_i,\bx_{i'})\in\R^d\times\R^d$. 
In other words, the interaction kernel $\Phi$ can be factorized (as composition of functions) as a {\bf{reduced interaction kernel}} $\phi:\R^{d'}\rightarrow\R$, and {\bf{reduced variables}} $\bvar:\R^d\times\R^d\rightarrow\R^{d'}$, i.e.
\begin{equation}
\Phi(\bx_i,\bx_{i'})=\phi(\bvar(\bx_i,\bx_{i'})).
\label{e:PhiFactorization}
\end{equation}
An important example is $\bvar(\bx_i,\bx_{i'}):=||\bx_i-\bx_{i'}||$, i.e. $\Phi$ is in fact just a function of pairwise distances, a one-dimensional variable.
Given a set of discrete observations of trajectory data $$\{\bx_i^{(m)}(t_l), \dot\bx_i^{(m)}(t_l)\}_{i, l, m = 1}^{N, L, M}\,,$$ with $0 = t_1 < \cdots < t_L = T$ and initial conditions $\bx_i^{(m)}(t_1)$ sampled i.i.d. from $\mu_0$ (some unknown probability distribution on $\R^{dN}$) for $m=1, \ldots, M$, we are interested in estimating the interaction kernel $\Phi$.
Note that this is not a regression problem, but an inverse problem, since values of the function $\Phi$ are not observed (only averages of such values, as per the right-hand side of \eqref{eq:first_order}, are).
In the works that considered this problem (\cite{BFHM2017,LZTM2019,MTZM2020}), {\em{the reduced variables $\bvar$ were given}}, and $\phi$ was estimated as a function of $\bvar$.
One of the main takeaways of those works is that the estimation of $\phi$ is not cursed by the dimension $dN$ of the state space of the system, nor by the dimension $2d$ of pairs of agent states, and, under suitable stable identifiability conditions, it is not harder than regression in $d'$ dimensions, with the min-max learning rate of $M^{-2/(2+d')}$ (up to log factors) being achievable, with efficient algorithms.

In this work, we consider the case where the reduced variables $\bvar$ are {\em{not}} known: both $\bvar$ and $\phi$ need to be estimated from observations. 
We wish to accomplish this while exploiting the factorization \eqref{e:PhiFactorization}, in order to obtain rates of estimations that are independent of both the dimension $dN$ of the state space and of the dimension of $2d$ of pairs of states, and only dependent on $d'$. 
We relate this problem to the classical multi-index model: one considers functions $f(\bx)=g(A\bx)$, where $\bx\in\R^D$, $A:\R^D\rightarrow\R^{d'}$ is a linear map, $g:\R^{d'}\rightarrow\R$, and given i.i.d. samples $(\bx_i,f(\bx_i))_{i=1}^n$, one constructs estimators of $g$ and $A$. We defer to \cite{Martion2021multi} for discussions, references, and a novel technique for estimating $(g, A)$ that avoids the curse of dimensionality and is computationally efficient.

In our setting, we cannot directly apply the techniques of multi-index regression, as we are facing an inverse problem.
We therefore consider a multi-task learning setting, where we are allowed to learn on systems with a varying number $N$ of agents, driven by interaction kernels sharing the same variables $\bvar$, and then transfer to other systems with different $N$ or $\phi$, but sharing the same $\bvar$.
Crucially, when $N=2$, it is possible to re-cast the estimation problem for $\phi$ and $\bvar$ as a multi-index regression problem, opening the way to efficient estimation.
Transferring to other systems with different $N$ is straightforward, and to systems with different $\phi$'s but the same variables $\bvar$ may be achieved by the techniques of \cite{LZTM2019,MTZM2020}.

\noindent{{\bf{Related Methods}}}.
There are many related approaches in learning the governing structures from observation of dynamical trajectory data, i.e. \cite{BCCCCGLOPPVZ2008, herbert2011inferring, BPK2016}; see also \cite{MTZM2020} for further references.  However, most of them do not take into consideration the special structure of self-organized dynamics.  Our learning method follows in the line of research of \cite{BFHM2017, LZTM2019}, which exploits both the structure of the equations, and the assumption that $\Phi$ depends on a small number of reduced variables $\bvar$ (e.g. pairwise distance).  The learning framework developed in \cite{BFHM2017} focused on the convergence case when $N$ goes to infinity, while \cite{LZTM2019} consider the convergence when $M$ goes to infinity; extensions include \cite{ZMM2020, MTZM2020, MMQZ2021}.  In these works, the reduced variables are known and the focus was on learning the interaction kernels; here we generalize the framework to the case when the variables $\bvar$ are unknown.
\section{Learning Method}\label{sec:learn}
We assume the discrete observations of trajectory data, $\{\bx_i^m(t_l), \dot\bx_i^m(t_l)\}_{i, m, l = 1}^{N, M, L}$ with  $0 = t_1 < \cdots < t_L = T$, is equi-spaced in time: $t_{l+1}-t_l=T/(L-1)$ for all $l$'s.  
We seek to infer the variables $\bvar$ and the reduced interaction kernel $\phi$ in the factorization in \eqref{e:PhiFactorization}.

We proceed by further assuming that $\bvar$ can be factorized as $\bvar(\bx_i,\bx_{i'}):=B\by(\bx_i,\bx_{i'})$, where $\by:\R^d\times\R^d\rightarrow\R^D$ is a {\em{known}} feature map to high-dimensional feature vectors, and the feature reduction map $B:\R^D\rightarrow\R^{d'}$ is an {\em{unknown}} linear map to be estimated. What really matters is the range of $B$, and without loss of generality we assume $B$ orthogonal. The factorization \eqref{e:PhiFactorization} becomes: $\Phi(\bx_i, \bx_{i'}) = \phi(B\by(\bx_i,\bx_{i'}))$; we estimate $\phi$ and $B$ in two steps.

\noindent{\bf{Step $1$. Estimating $B$ and ${\bvar}$}}.
The techniques of \cite{Martion2021multi}, for estimation of the multi-index model, are only applicable in the regression setting. However, we note that if we allow the learning procedure to conduct ``experiments'' and have access to observations of \eqref{eq:first_order} for different values of $N$, and in particular for $N=2$, then from those observations it can extract equations that reveal the values taken by the interaction kernels, taking us back to a regression setting. We then apply the approach of \cite{Martion2021multi} to obtain $\hat B$, estimating $B$.  
This technique is not cursed by the dimension $D$, making it possible to increase the range of the feature map $\by$ with relatively small additional sampling requirements.
While the choice of the feature map $\by$ is typically application-dependent, and can incorporate symmetries, or physical constraints on the system, a rather canonical choice is the map to polynomials in the states, and  here we restrict ourselves to second-order polynomials, and therefore assume that: 
\begin{equation}\label{e:yfeatureMap}
\begin{aligned}
\by (\bx_i,\bx_{i'}) := \big[ \bx_i , \bx_{i'} , &( (\bx_i)_j(\bx_i)_{j'})_{\substack{j\le j'}}, (( {\bx_{i'})_j}(\bx_{i'})_{j'})_{\substack{j\le j'}}, \\ 
& ((\bx_i)_{j} (\bx_{i'})_{j'})_{\substack{j,j'=1,\dots,d}} \big]^T \!\!\!\in \R^{D}\,,
\end{aligned}
\end{equation}
with $D=2d^2 + 3d$.  Of course higher order terms, beyond the quadratic ones, could be added to the feature map $\by$.  

\noindent{\bf{Step $2$. Estimating $\phi$}}.
Once an estimate $\hat B$ for the feature reduction map $B$ has been constructed, we proceed to estimate $\phi$. 
We project the pairs of states, using $\hat B$, to $\R^{d'}$, and use the non-parametric estimation techniques of \cite{LZTM2019}: we minimize, over functions $\psi$ in a suitable hypothesis space $\hypspace$, the error functional
\[
\begin{aligned}
\mE(\psi)\!\!:=\!\!\frac{1}{NLM}\!\!\sum_{i, l, m = 1}^{N, L, M}\bigg\|{\dot\bx_{i, l}^m\! - \!\frac{1}{N}\!\sum_{\substack{i' = 1}}^N\psi(\hat{B}\by_{i, i', l}^m)(\bx_{i', l}^m - \bx_{i, l}^m)}\bigg\|^2
\end{aligned}
\]
where $\bx_{i, l}^m := \bx_i^{(m)}(t_l)$ for $i = 1, 2$, and $\by_{i, i', l}^m := \by_{i, i'}^{(m)}(t_l)$ for $i, i' = 1, 2$, $i \neq i'$. $\hypspace$ is a convex and compact (in the $L^\infty$ norm) set of functions of the estimated variables $\hat B\by$, e.g. spanned by splines with knots on a grid, or piecewise polynomials. As in classical nonparametric estimation, the dimension of $\hypspace$ will be chosen as a suitably increasing function of the number of training trajectories $M$.
\subsection{Details on estimating the reduced variables and kernel}
Step $1$ described above requires two key ideas: first of all, mapping the inverse problem of estimating the interaction kernel to a regression problem, by considering the problem for $N=2$ (from which the solution is immediately transferred to any $N\ge2$). Secondly, given input-output pairs $((\bx_i^m(t_l),\bx_{i'}^m(t_l)), \phi(B\by(\bx_i^m(t_l),\bx_{i'}^m(t_l))))$, with $\phi$ and $B$ unknown and $\by$ known, we estimate $B$ while trying to avoid the curse of dimensionality -- meaning both $2d$ (the dimension of $(\bx_i,\bx_{i'})$, the domain of $\Phi$), and $D$ (the dimension of the domain of $B$, which is the range of $\by$).

\vskip0.1cm
\noindent{\bf{First-order system with $N=2$ agents}}.
We consider the controlled experiments where we observe a first-order system of only $N=2$ agents (we drop the dependence on $t$ for now to ease the notation), i.e.,
\[
\begin{aligned}
\dot\bx_1 = \frac{1}{2}\Phi(\bx_1, \bx_2)(\bx_2 - \bx_1)\,,\,\,
\dot\bx_2 = \frac{1}{2}\Phi(\bx_2, \bx_1)(\bx_1 - \bx_2). \\
\end{aligned}
\]
We can transform the equations to
\[
\begin{aligned}
\frac{2\inp{\dot\bx_1, \bx_2 - \bx_1}}{\norm{\bx_2 - \bx_1}^2} = \Phi(\bx_1, \bx_2)\,,\, \frac{2\inp{\dot\bx_2, \bx_1 - \bx_2}}{\norm{\bx_1 - \bx_2}^2} = \Phi(\bx_2, \bx_1). \\
\end{aligned}
\]
Let $\by_{1,2}:=\by(\bx_1,\bx_ 2) \in \R^{D}$ as in \eqref{e:yfeatureMap}; we obtain a regression problem:
\[
\phi(B\by_{1, 2}) = z_{1, 2} \coloneqq 2\frac{\inp{\dot\bx_1, \bx_2 - \bx_1}}{\norm{\bx_2 - \bx_1}^2} = \Phi(\bx_1,\bx_2). \\
\]
Similar definitions are used for $\by_{2, 1}$ and $z_{2, 1}$.  We re-index the observations as $\{\by_q\}_{q=1}^Q:=\{\by_{1, 2}^{(m)}(t_l), \by_{2, 1}^{(m)}(t_l)\}_{l, m = 1}^{L, M}$ and the corresponding $\{\bz_q\}_{q=1}^Q:=\{z_{1, 2}^{(m)}(t_l), z_{2, 1}^{(m)}(t_l)\}_{l, m = 1}^{L, M}$, with $Q=2LM$. We are now ready to estimate $B\in \R^{d' \times  D}$.
\vskip0.1cm
\noindent{\bf{Multiplicatively Perturbed Least Squares}} (MPLS) is a computationally efficient algorithm introduced in \cite{Martion2021multi} for estimating multi-index models, which we apply here to the estimation of $B$. It decomposes the regression function into linear and nonlinear components: $\Phi(\bx_{i}, \bx_{i'}) = \phi(B\by_{i, i'}) = \inp{\beta, \by_{i, i'}} + g(A\by_{i, i'})$, where $(i, i') = \{(1,2), (2, 1)\}$ and $g$ is orthogonal to linear polynomials.  The intrinsic domain of $\Phi$, the row-space of $B$, is spanned by $\beta$ and the rows of $A$, estimated as follows:

\textbf{Algorithm [MPLS]} 
Inputs: training data $\{\by_q, z_q\}_{q = 1}^Q$, 
partitioned into subsets $\mathcal{S}$ and $\mathcal{S}'$ with sizes $|\mathcal S|, |\mathcal{S}'|\ge\lfloor {Q}/{2}\rfloor$; 
parameters $K$ and $\lambda$, with $K\gtrsim d'\log d'$ and $\lambda \approx {1}/{D}$ (choosing $K$ and $\lambda$ is discussed in \cite{Martion2021multi}).
\begin{enumerate}
\item Estimate $\beta$ via ordinary least squares on $\mathcal{S}'$:
\[ 
\begin{aligned}
\hat\beta \coloneqq \argmin{\beta'\in\mathbb{R}^{D}}\frac{2}{|\mathcal{S}'|} \sum_{(\by', z')\in\mathcal{S}'}\Big(z' - \inp{\mathbf{\beta'}, \by'}\Big)^2.
\end{aligned}
\]
\item Let $\mathcal{R} \coloneqq \Big\{(\by - \langle \hat\beta, \by\rangle \|\hat\beta\|^{-2} \hat \beta, z - \inp{\hat\beta, \by}) \big| (\by, z)\in\mathcal{S}\Big\}$ be the residual data of this approximation on $\mathcal{S}$, paired with $\by$s projected on $\hat\beta^\perp$.
\item Pick $\bu_1,\ldots,\bu_K$ in $\R^{D}$ (e.g., a random subset of $\{\by_q\}_{q = 1}^Q$). For each $\bu_i$, center the residuals to their weighted mean: with $w(\tilde \by,\bu_i)=\exp(-\lambda\norm{\tilde \by - \bu_i}^2)$,
{\small
\[
\widetilde{ \mathcal{R}} \coloneqq \Bigg\{\bigg({\tilde \by, r - \frac{\sum_{(\tilde \by, r) \in \mathcal{R}} w(\tilde \by; \bu_i)r}{\sum_{(\tilde \by, r) \in \mathcal{R}} w(\tilde \by;\bu_i)}}\bigg)\bigg| (\tilde \by, r) \in \mathcal{R}\Bigg\} \,,
\]}
\!\!\!and compute the slope perturbations via least squares:
\[  
\mathbf{\hat p}_i \coloneqq \argmin{\mathbf{p} \in \R^{D}} \frac{2}{|\mathcal{S}|} \sum_{(\tilde \by, \tilde{r}) \in \widetilde{\mathcal{R}}} (w_k(\tilde \by, \bu_i)\tilde{r} - \inp{\mathbf{p}, \tilde \by})^2\,.
\]
\item Let $\hat P\in\R^{K \times D}$ have rows $\mathbf{\hat p}_i$, $i=1,\dots,K$, and compute the rank-$d'$ singular value decomposition of $\hat P \approx U_{d'} \Sigma_{d'} V_{d'}^T$. Return $\hat A := V_{d'}^T \in\mathbb{R}^{d'\times D}$ and $\hat\beta$.
\end{enumerate}
The consistency of the estimators and convergence rates are discussed in \cite{Martion2021multi}; a key outcome of that analysis is that the sampling requirements for accurately estimating $\beta$ and $A$ are not exponential in $D$, but a low-degree polynomial in $D$, therefore avoiding the curse of dimensionality. Those results hold when the training data is sampled i.i.d. from some distribution over $\by$, while here the samples are clearly not independent along each trajectory, but they are on the $M$ different trajectories, as the initial conditions are sampled independently. 
It remains to convert the computed $\hat A \in \mathbb{R}^{d'\times D}$ and $\hat \beta \in \mathbb{R}^{1\times D}$ (from linear components) into an estimate of the low-dimensional space $B \in \mathbb{R}^{d'\times D}$. This is done by computing the regression error of a 1-dimensional constant spline approximation to $\{\langle \by_q, \nu\rangle, z_q\}_{q=1}^Q$ for $\nu\in\{\hat \beta, \hat A_{d'}\}$; comparing this error for $\hat \beta$ and the $d'$-th singular vector of $\hat P$. If the former's error is less than the latter's, which suggests that the linear component is a more significant feature, then we take $\hat B$ to be the row vector $\hat\beta / \|\hat\beta\|$ stacked on the first $d' -1$ rows of $\hat A$, otherwise we let $\hat B \coloneqq \hat A$. Note that because the $\tilde\by$s used to compute $\hat A$ are projected away from $\hat \beta$, $\hat B$ will have orthonormal rows in either case. This procedure could be done at the beginning to identify useful vectors on the unit ball in $D$ dimensions (as in \cite{JLT2009}), however this would involve checking a number of vectors exponential in $D$, and is not computationally feasible here. MPLS allows us to the reduce the problem to one comparison, regardless of $D$.
%
\subsection{Performance Measures}
To measure the accuracy in approximating the feature reduction map $B$, we use
$\text{Err}_{B} = ||{B^{\top}B - \hat{B}^{\top}\hat{B}}||$.
To measure how well $(\hat{B}, \hat\phi)$ approximates $(B, \phi)$ we use
\begin{equation}\label{eq:phi_error}
\begin{aligned}
\text{Err}_{\phi}^2 := 
\int_{\by \in \R^D} \abs{\phi(B\by) - \hat\phi(\hat{B}\by)}^2W^2(\by) \, d\rho_T(\by)\,,
\end{aligned}
\end{equation}
where $W(\by_{i, i'}) = \norm{\bx_i - \bx_{i'}}$; its relative variant is
${\text{Err}}^{\text{rel}}_{\phi} := {\text{Err}_{\phi}}/({\int_{\by \in \mathcal{Y}} \abs{\phi(B\by)}^2W^2(\by) \, d\rho_T(\by)})^{1/2}$.
The dynamical system, together with its random initial conditions sampled from $\mu_0$ on $\R^{2d}$, induces the probability measure $\rho_T(\by)$ on $\by \in \R^{D}$, given by
$
\rho_T(\by) = \E_{(\bx_1(t_1),\bx_2(t_1)) \sim \mu_0}\int_{t = 0}^T\sum_{{i, i' = 1, i \neq i'}}^N\delta_{\by_{i, i'}(t)}(\by)\,.
$

We measure the trajectory accuracy between the observed trajectory, $\{\bx_{i}(t)\}_{i = 1}^N$, and the trajectory $\{\hat\bx_i(t)\}_{i = 1}^N$ obtained by the system driven by $(\hat{B}, \hat\phi)$ (over matching initial conditions) as
\begin{equation}\label{eq:traj_error}
\text{Err}_{\text{traj}}^2 := \E_{\mu_0}\frac{1}{NT}\sum_{i = 1}^N\int_{t = 0}^T\norm{\bx_i(t) - \hat\bx_i(t)}^2\, dt\,,
\end{equation}
with the expectation taken over initial conditions $(\bx_i(t_1))_{i=1}^N$ sampled i.i.d. from $\mu_0$.  
A relative version ${\text{Err}}^{\text{rel}}_{\text{traj}}$, so that trajectory errors from different dynamics are more comparable, is obtained by dividing $\text{Err}_{\text{traj}}$ by
$$(\E_{\bx_i(t_1)}[\frac{1}{NT}\sum_{i = 1}^N\int_{t = 0}^T\norm{\bx_i(t)}^2\, dt])^{1/2}\,.$$
For a set of trajectories with $M$ different initial conditions, we report the trajectory error as $\text{Err}_{\text{traj}, \text{mean}}^{\text{rel}}$.
\subsection{Convergence Characteristics and Limitations}
We expect our approach to preserve the strong convergence properties of MPLS and  of the estimators of interaction kernels in \cite{MTZM2020}, in particular avoiding the curse of dimensionality in $Nd$ and $d^2$.
Limitations of our approach include the need of available training data for $N = 2$, and the knowledge of the feature map $\by$, such that $\xi$ is a linear function of $\by$. 
\section{Numerical Examples}\label{sec:numerics}
\begin{figure}[t]
\begin{center}
\includegraphics[width =4.0cm]{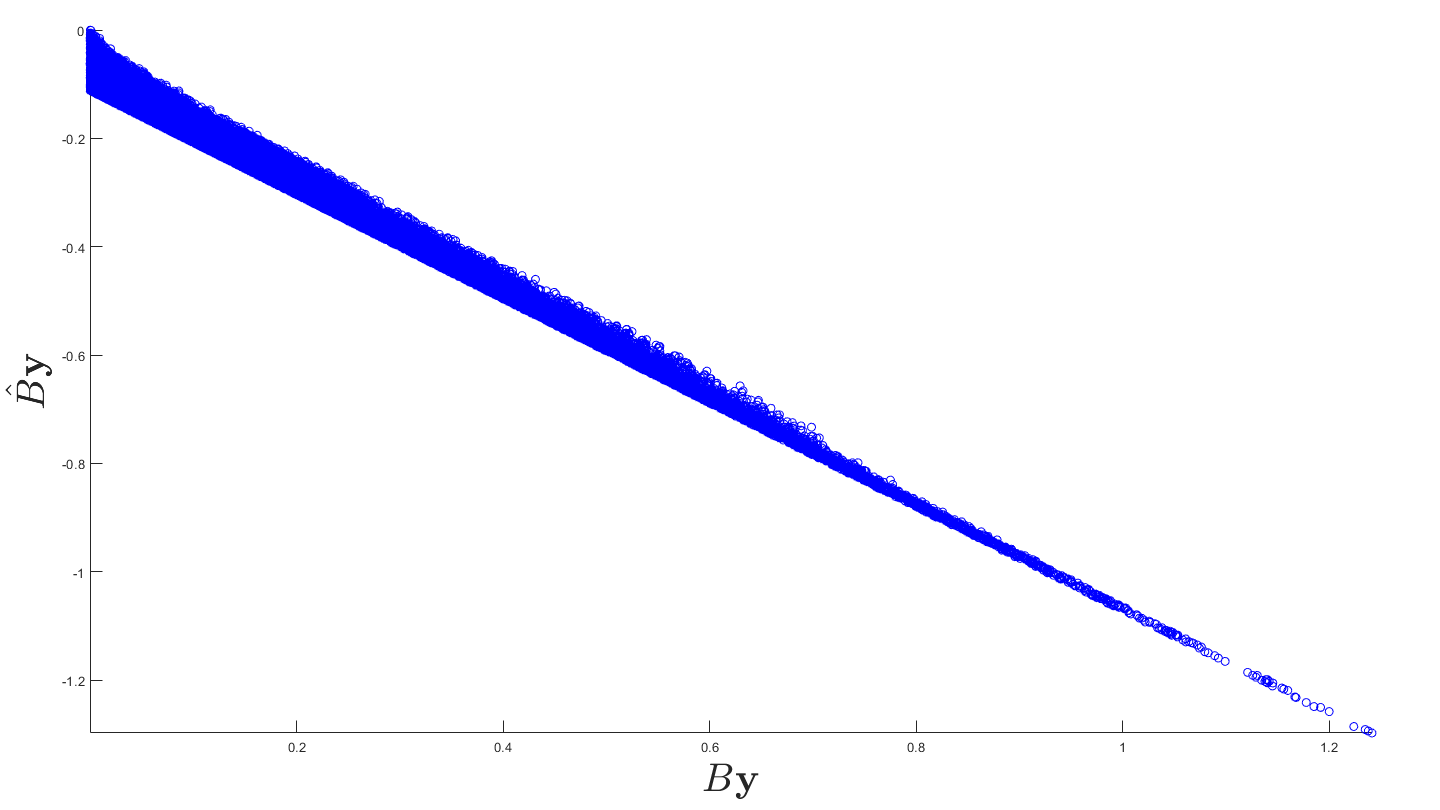} 
\,
\includegraphics[width =4.0cm]{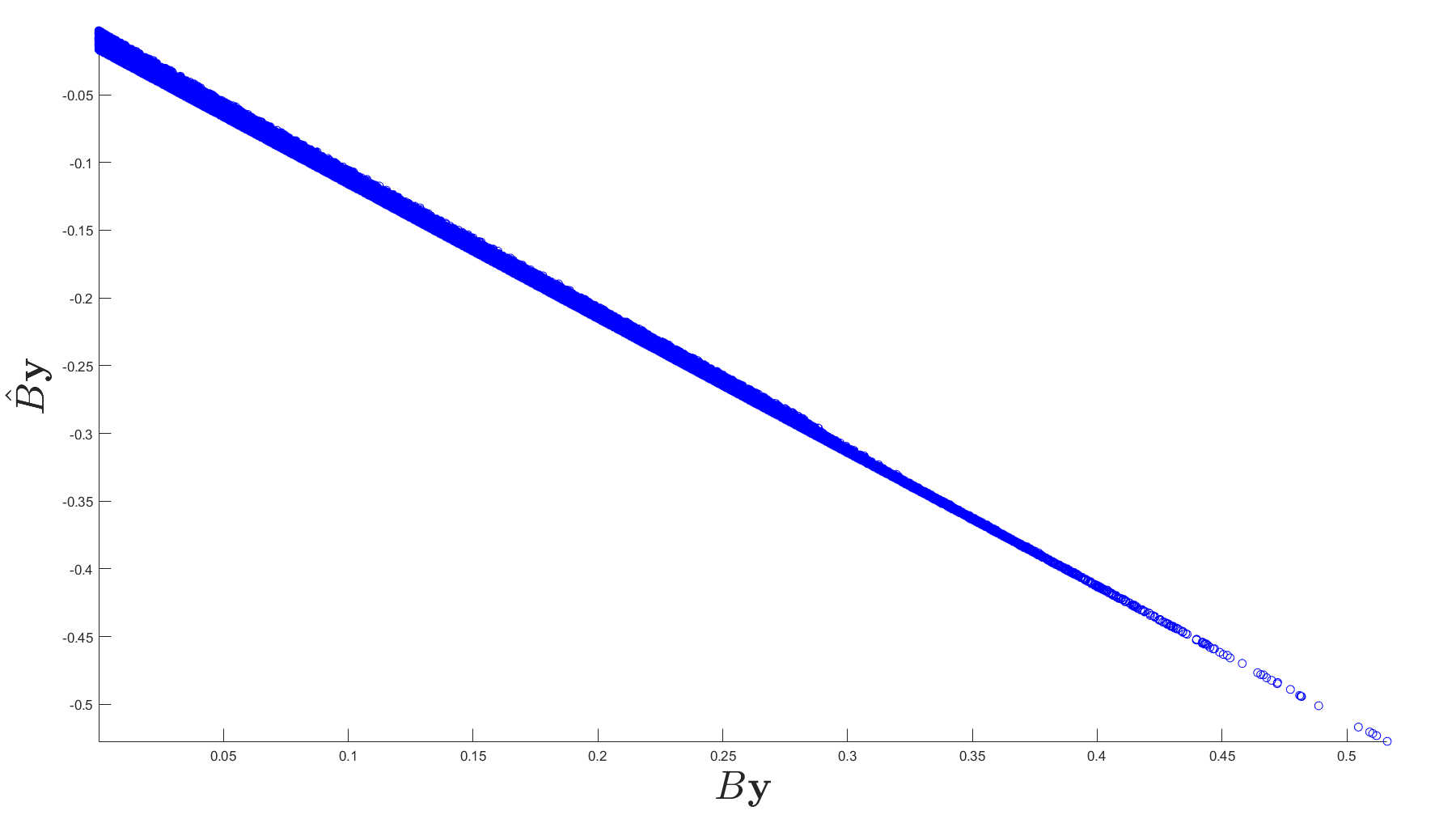} 
\caption{$B\by$ vs. $\hat{B}\by$. Opinion Dynamics on the left and PowerLaw Dynamics on the right.}
\label{fig:FM_comp}
\end{center}
\end{figure}

\begin{figure}[hb]
\begin{center}
\includegraphics[width=8cm]{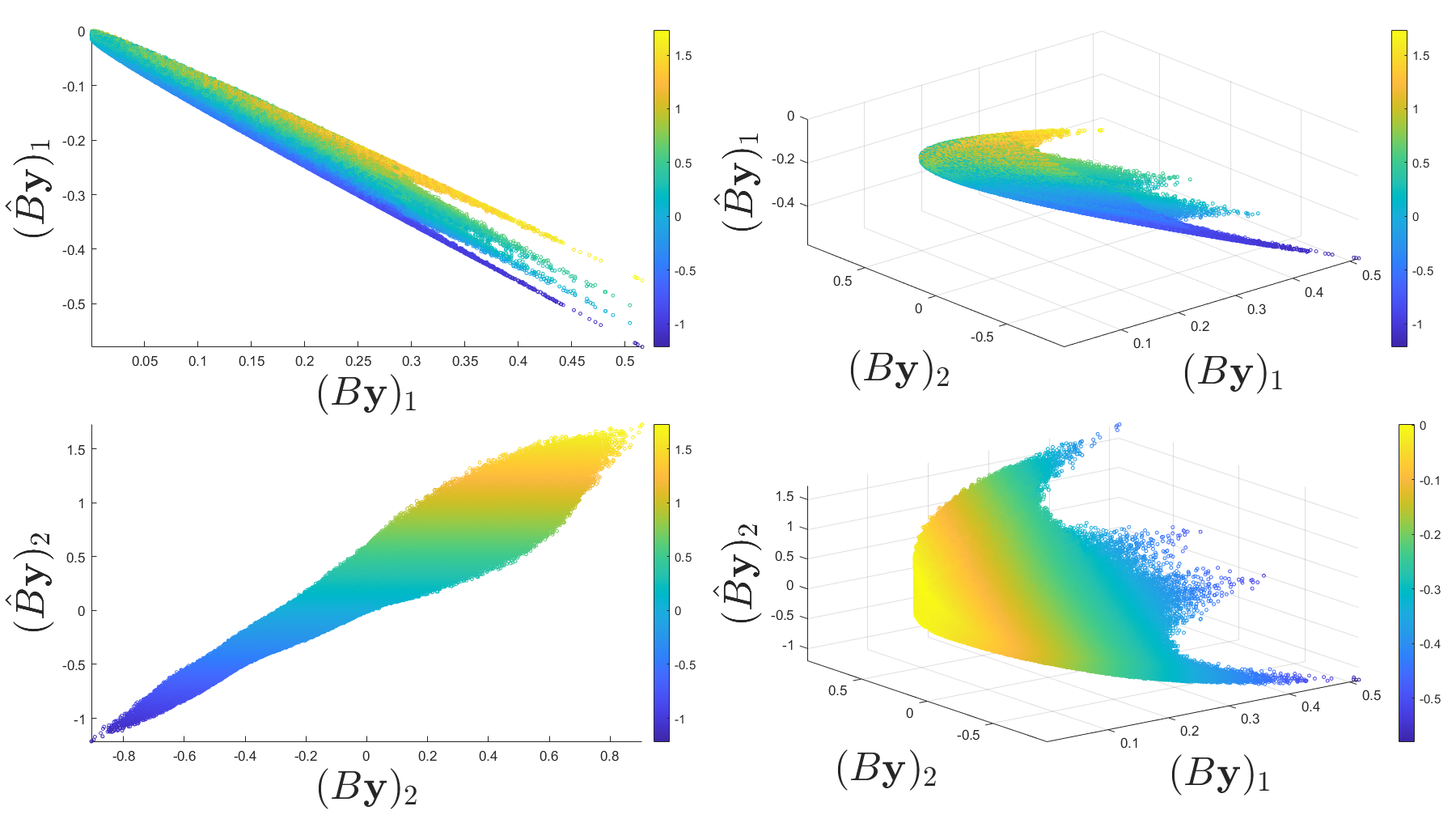}    
\includegraphics[width=8cm]{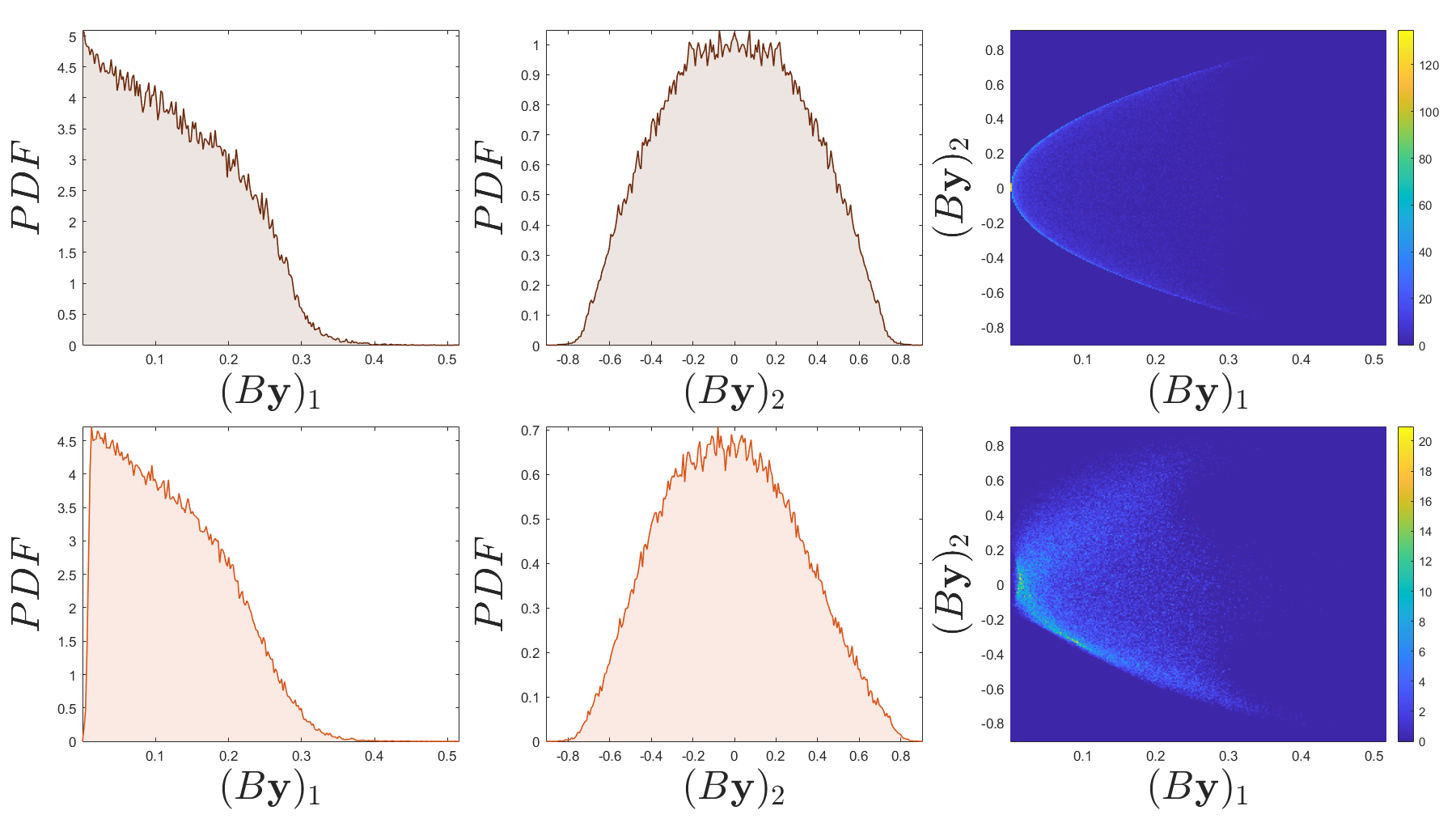}    
\caption{
{\em{Top}}: The smooth bijective relationship between $B\by$ and $\hat{B}\by$: in the first row the colors reflect the values of $(\hat{B}\by)_2$; in the second row they reflect those of $(\hat{B}\by)_1$.
{\em{Bottom}}: Marginal and joint distributions of $B\by$ (first row) is well-matched by that of $\hat{B}\by$ (second row), here on the same $(B\by)_1$, $(B\by)_2$-axis, respectively, for ease of comparison.} 
\label{fig:PLwDC_FM_dist_comp}
\end{center}
\end{figure}

We consider three kinds of dynamical systems exhibiting different emergent patterns: opinion dynamics, where clustering of agents emerges and the interaction kernels are compactly supported and discontinuous; power-law dynamics, where ring-like patterns emerge and the interaction kernels are powers of the pairwise distance variables; a modified power-law dynamics where a directional influence is added to steer the agents into a preferred direction.  The shared parameters of these experiments are given in Table \ref{tab:comm_param}.
\begin{table}[H]
\begin{center}
\caption{Common Parameters}\label{tab:comm_param}
\begin{tabular}{c | c | c | c | c | c | c} 
$M$ & $M_{\text{transfer}}$ & $N$ & $L$ & $T$ & $d$ & $D$ \\
\hline
$50000$ & $500$ & $2$ & $5$ & $1$ & $2$ & $14$\\
\end{tabular}
\end{center}
\vskip-0.5cm
\end{table}
Here $M_{\text{transfer}}$ is used to generate the initial conditions for $N_{\text{transfer}} = 20 \gg N = 2$,  and it is significantly smaller than $M$, used for $N=2$ to learn $\bvar$, showcasing an excellent transfer learning capability.  
For the $N = 2$ case, we have the mapping $\by(\bx_i, \bx_{i'})$ given by \eqref{e:yfeatureMap}.
Fig. \ref{fig:FM_comp} compares the reduced features $B\by$ and $\hat{B}\by$.
Notes that $\hat{B}\by$ estimates $B\by$ up to orthogonal transformations; this effect can be seen in Fig. \ref{fig:FM_comp} and Fig. \ref{fig:PLwDC_FM_dist_comp}. However, when we visually compare the interaction kernels and the distributions of $B\by$ and $\hat{B}\by$, we will coordinate the plots on the same $B\by$-axis, for the purpose for easier visual comparison.

After we obtain $\hat{B}$ (or, in the Oracle case, when $B$ is given), we estimate $\phi$, $\hat\phi_1$, from the data $\hat{B}\by$ (or $\hat\phi_2$, for the data $B\by$) using the following steps: $1)$ estimate the support from $\hat{B}\by$ (or $B\by$); $2)$ build a basis on a uniform partition of the support with the optimal number of basis functions $n = ({LM}/{\log(LM)})^{{d'}/{(d' + 2)}}$ using either piecewise polynomials or clamped B-splines as basis functions; $3)$ solve for $\hat\phi_1$ or $\hat\phi_2$ as a linear combination of the basis functions (see \cite{LZTM2019}).  We run learning trials over $10$ training sets in order to demonstrate the effectiveness of the learning results as the training data is randomly generated, and report the mean and standard deviation of the performance estimators over these trials.

\begin{figure}[t]
\begin{center}
\includegraphics[width=8.4cm]{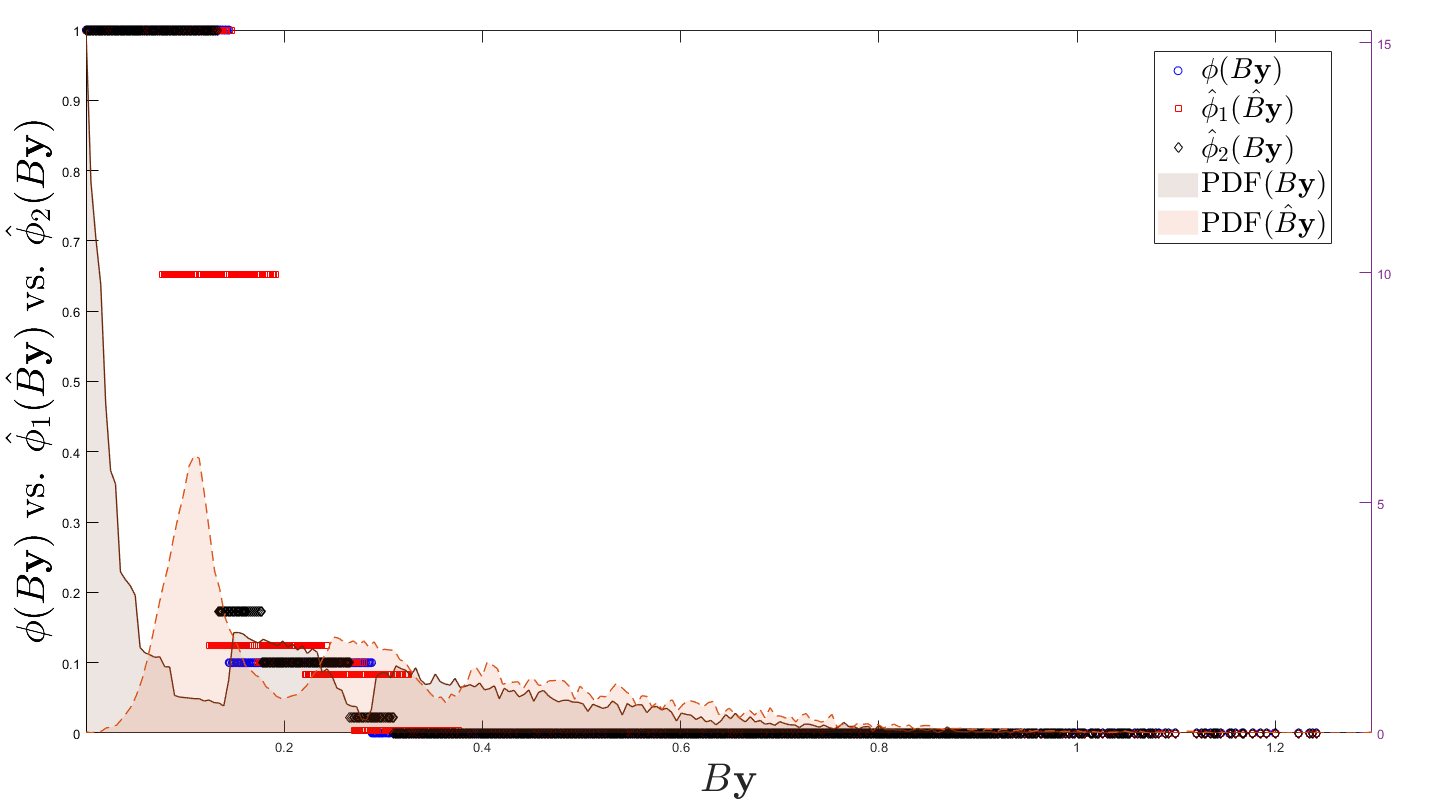}    
\caption{Comparing estimated interaction kernels vs. the truth, and the distribution of $\hat{B}\by$ and $B\by$. is shown in light/dark red, respectively.  For ease of comparison, all of the plots are on the same $B\by$-axis. As shown in the figure, our learned feature variable and $\phi$ provide a faithful approximation to the original, with the exception of not detecting the point of discontinuity close enough; however, even with known feature variable, the learned interaction still has trouble approximating the function at discontinuity.} 
\label{fig:OD_phi_comp}
\end{center}
\end{figure}

\begin{figure}[hb]
\begin{center}
\includegraphics[width=8.4cm]{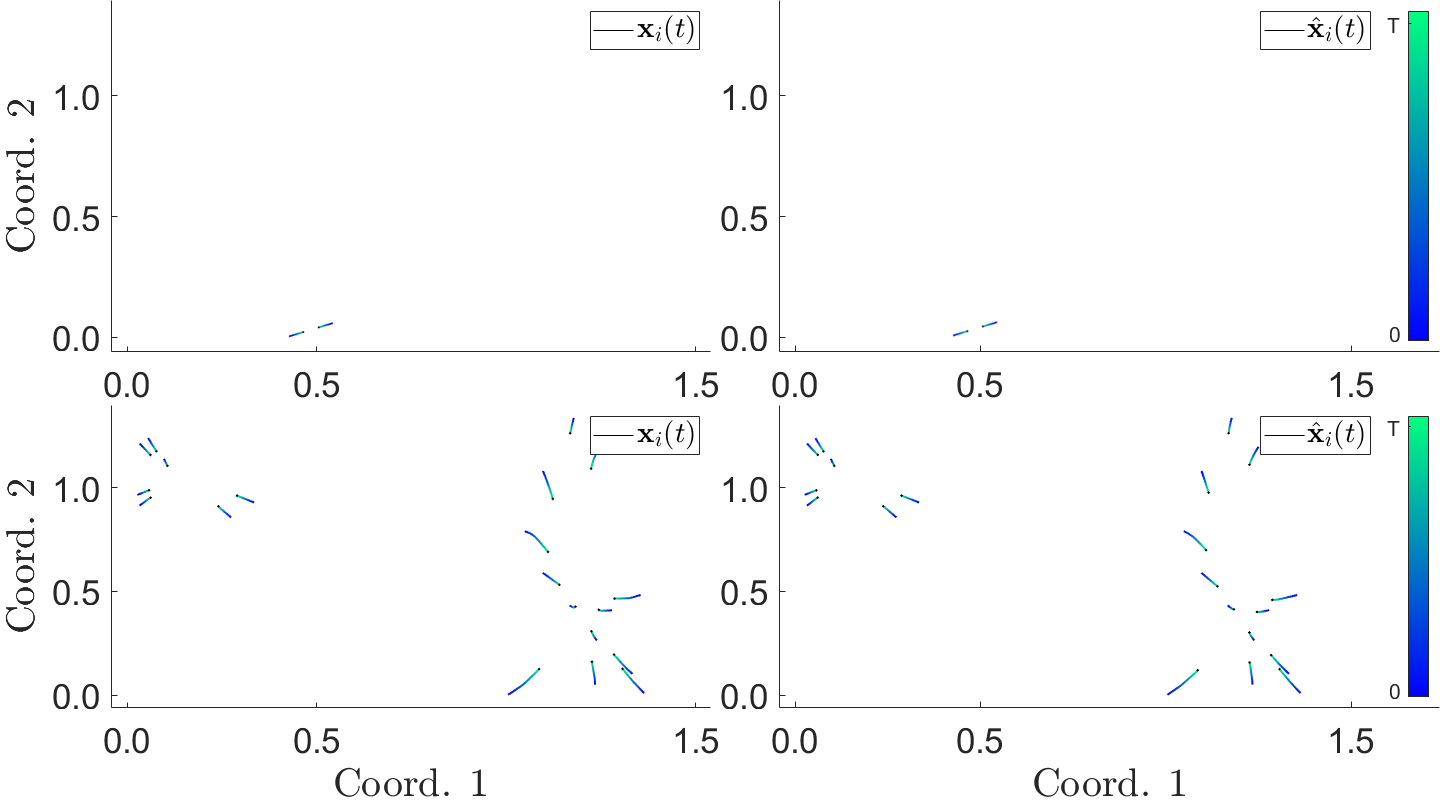}    
\caption{$\{\bx_i(t)\}_{i = 1}^N$ vs. $\{\hat\bx_i(t)\}_{i = 1}^N$.  The first row show the comparison of the trajectories over a new random initial conditions not from the training data set ($N=2$).  The second row shows the comparison of the trajectories over a new initial condition ($N=20$).  Our learned systems can provide a reliable reproduction of the dynamics, quantified in Table \ref{table:OD_err}.}
\label{fig:OD_traj}
\end{center}
\end{figure}

\noindent{\bf{Opinion Dynamics (OD)}}.
We consider the system of \cite{MT2014}, where $\Phi(\bx_i, \bx_{i'}) = \phi(B\by(\bx_i, \bx_{i'}))$, $B\by(\bx_i, \bx_{i'}) = {\norm{\bx_i - \bx_{i'}}^2}/({2\sqrt{3}})$ and
\[
\phi(\xi) = 0.1\cdot\mathbf{1}_{[0,\frac{1}{4\sqrt{3}})}(\xi)+1\cdot\mathbf{1}_{[\frac{1}{4\sqrt{3}},\frac{1}{2\sqrt{3}})}(\xi)\,,
\]
where $\mathbf{1}_I$ is the indicator function of set $I$, $\mu_0$ is uniform on $[0, 5]^2$, and $n = 28$. Table \ref{table:OD_err} shows the estimation errors for this system, where  $\text{Err}_{\phi}^{\text{rel}}$ (Oracle) indicates that the $L^2(\rho_T)$ error is calculated when the true $B$ is known. In particular, the proposed method has a relative accuracy in estimating $\phi$ comparable to the case when $B$ is known (Oracle case). We can obtain a faithful approximation of $B$ (in terms of $\hat{B}$) and of the interaction kernel using the pair $(\hat{B}, \hat\phi_1)$, with $2$-digit relative accuracy.
\begin{table}[H]
\begin{center}
\caption{OD Errors}\label{table:OD_err}
\scalebox{0.9}{
\begin{tabular}{l || l} 
$\text{Err}_{B}$ & $1.62 \cdot 10^{-1} \pm 5.7 \cdot 10^{-3}$ \\
\hline
$\text{Err}_{\phi}^{\text{rel}}$ & $3.07 \cdot 10^{-1} \pm 5.5 \cdot 10^{-3}$ \\
\hline
$\text{Err}_{\phi}^{\text{rel}}$ (Oracle) & $1.7 \cdot 10^{-1} \pm 3.7 \cdot 10^{-2}$ \\
\hline
$\text{Err}_{\text{traj}, \text{mean}}^{\text{rel}}$ & $8.1 \cdot 10^{-3} \pm 5.0 \cdot 10^{-4}$ \\
\hline
$\text{Err}_{\text{traj}, \text{mean}}^{\text{rel}}$ (Transfer) & $2.29\cdot 10^{-2} \pm 9.3 \cdot 10^{-4}$ \\
\end{tabular}
}
\end{center}
\vskip-0.5cm
\end{table}
We show the comparison of the two estimators (one using the feature reduction map $\hat{B}$ and the other using $B$) to the true interaction kernel in Fig. \ref{fig:OD_phi_comp}. A comparison of selected trajectories is shown in Fig. \ref{fig:OD_traj}.

\begin{figure}[hb]
\begin{center}
\includegraphics[width=8.4cm]{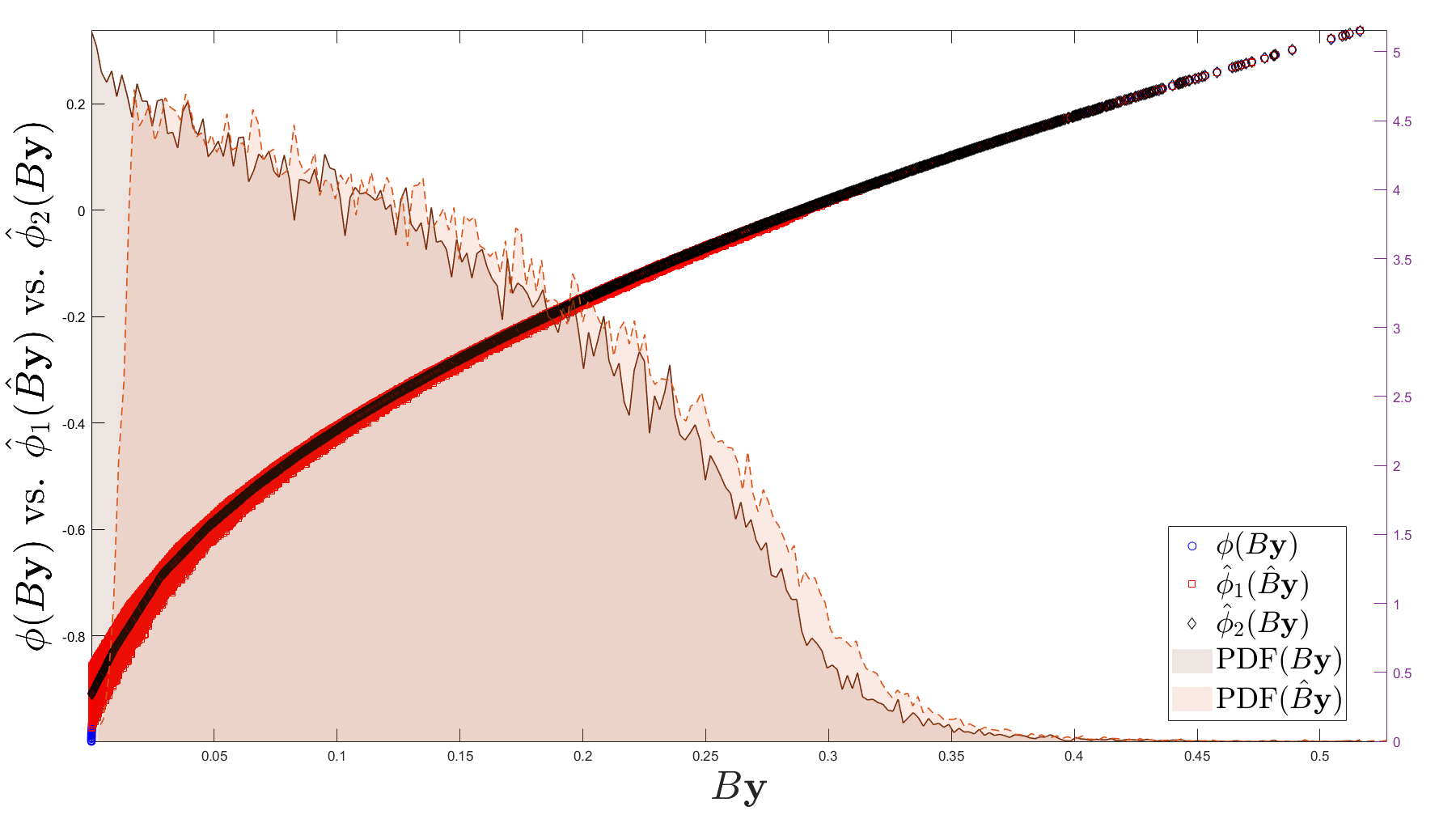}    
\includegraphics[width=8.4cm]{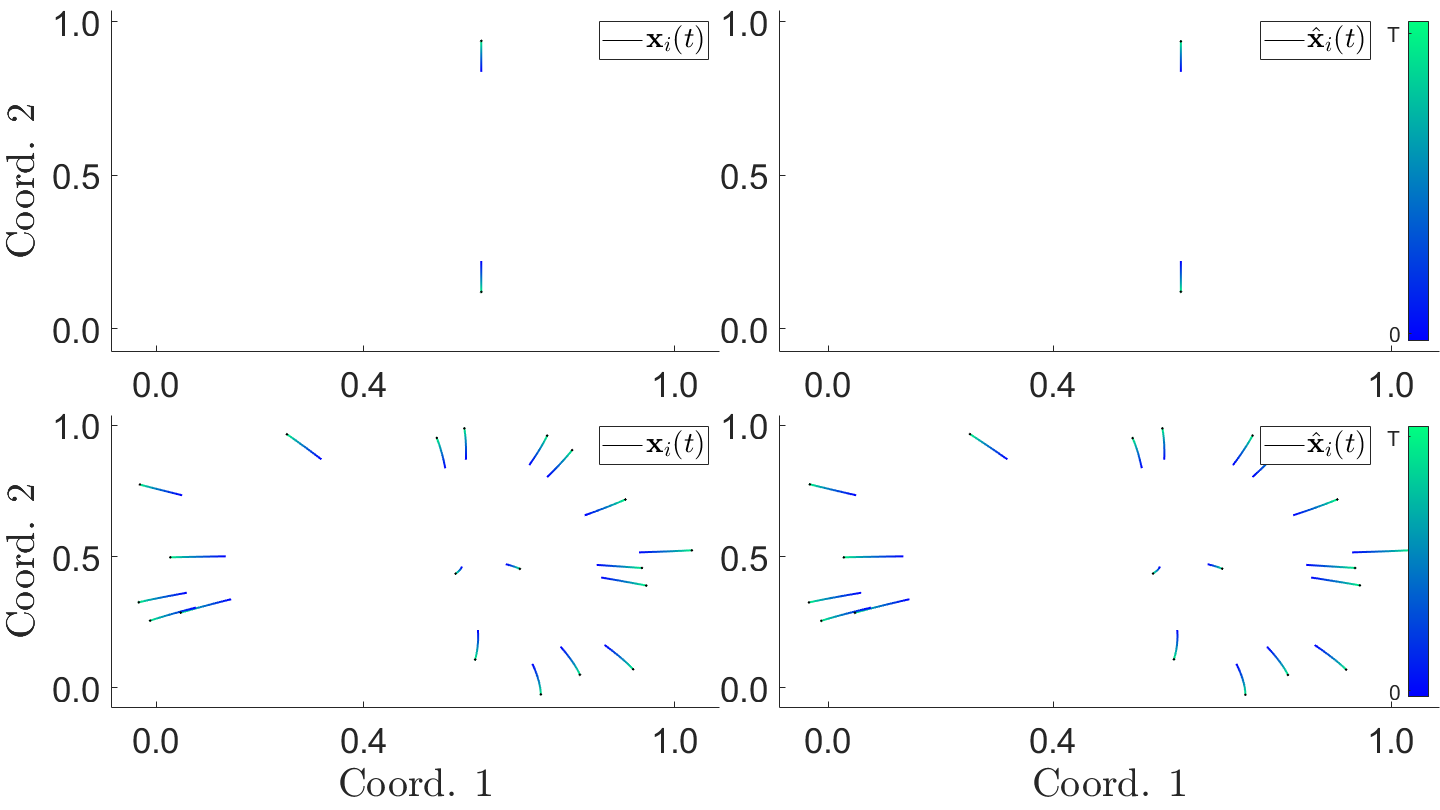}    
\caption{
{\em{Top}}: Estimated interaction kernels vs. the true one, and the distribution of $\hat{B}\by$ and $B\by$ (for ease of comparison, all the plots are on the same $B\by$-axis).
{\em{Bottom}}: $\{\bx_i(t)\}_{i = 1}^N$ vs. $\{\hat\bx_i(t)\}_{i = 1}^N$.  Similar layout as in Fig. \ref{fig:OD_traj}.  Our estimated system provides a reliable reproduction of the dynamics, quantified in Table \ref{table:PL_err}.}
\label{fig:PL_phi_traj_comp}
\end{center}
\end{figure}

\noindent{\bf{Power Law Dynamics (PL)}}.
We consider the dynamics in \cite{KSUB2011} where $\Phi(\bx_i, \bx_{i'}) = \phi(B\by(\bx_i, \bx_{i'}))$ with $B\by(\bx_i, \bx_{i'}) = {\norm{\bx_i - \bx_{i'}}^2}/{(2\sqrt{3})}$ and $\phi(\xi) = \sqrt{2\sqrt{3}\xi} - 1$.  $\mu_0$ is uniform on $[0, 1]^2$ and $n = 28$.  Table \ref{table:PL_err} shows the estimation errors, which are about $2$-digit relative accuracy in estimating the feature map and $\phi$, and lower for the trajectory errors.  
We compare the two estimators (using $\hat{B}$ and $B$ respectively) to $\phi$ as well as the corresponding dynamics in Fig. \ref{fig:PL_phi_traj_comp}.
\begin{table}[H]
\begin{center}
\caption{PL Errors}\label{table:PL_err}
\scalebox{0.9}{
\begin{tabular}{l || l} 
$\text{Err}_{B}$ & $3.7 \cdot 10^{-2} \pm 1.6 \cdot 10^{-3}$ \\
\hline
$\text{Err}_{\phi}^{\text{rel}}$ & $1.57 \cdot 10^{-2} \pm 6.3 \cdot 10^{-4}$ \\
\hline
$\text{Err}_{\phi}^{\text{rel}}$ (Oracle) & $2.0 \cdot 10^{-3} \pm 1.5 \cdot 10^{-4}$ \\
\hline
$\text{Err}_{\text{traj}, \text{mean}}^{\text{rel}}$ & $1.56\cdot 10^{-3} \pm 5.2 \cdot 10^{-5}$ \\
\hline
$\text{Err}_{\text{traj}, \text{mean}}^{\text{rel}}$ (Transfer) & $1.84\cdot 10^{-3} \pm 7.5 \cdot 10^{-5}$ \\
\end{tabular}
}
\end{center}
\vskip-0.5cm
\end{table}
For this continuous kernel, we are able to provide a learned pair $(\hat{B}, \hat\phi_1)$ with higher accuracy (when compared to the case of learning from opinion dynamics).

\begin{figure}[ht]
\begin{center}
\includegraphics[width=8cm]{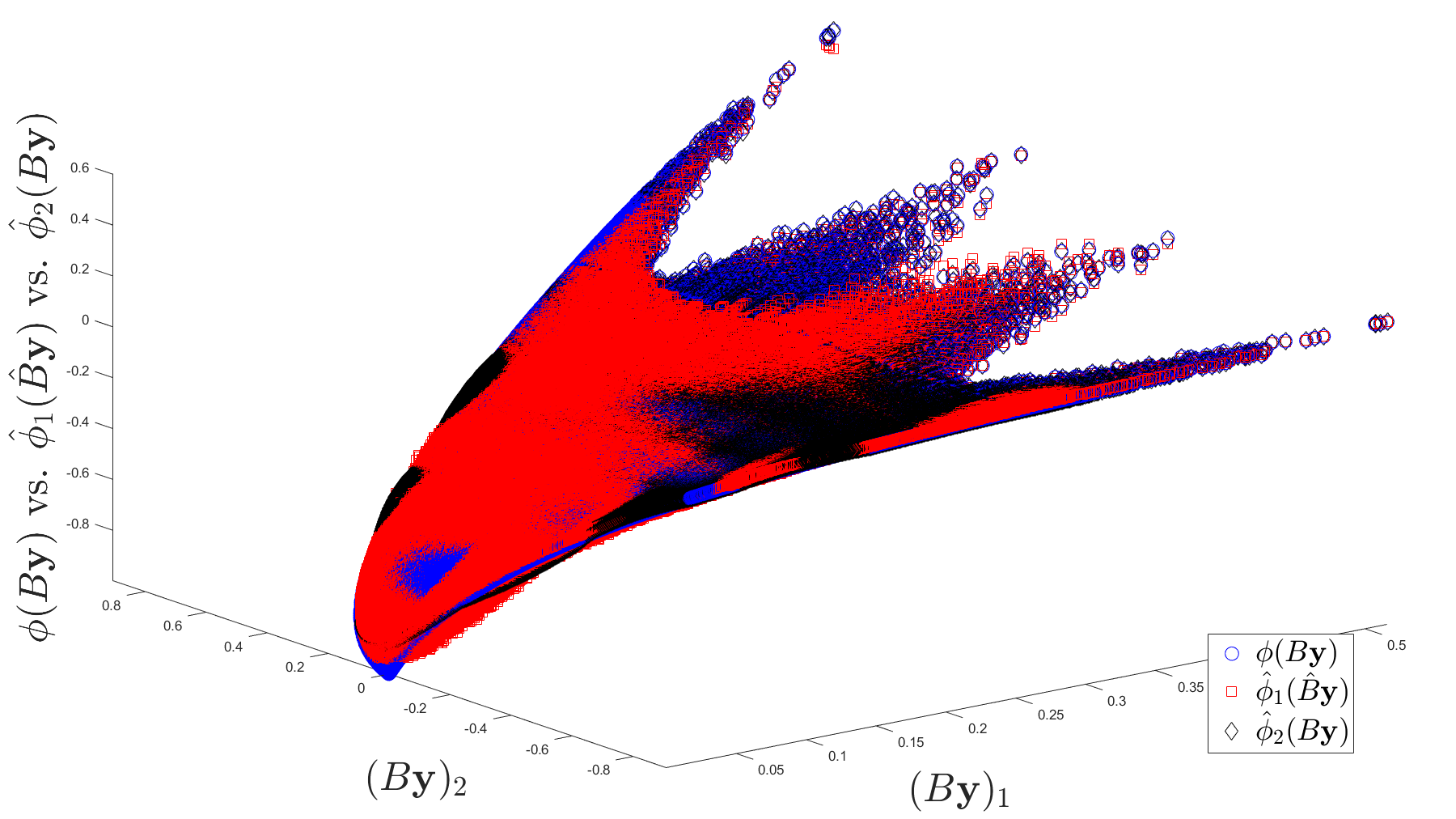}    
\includegraphics[width=8cm]{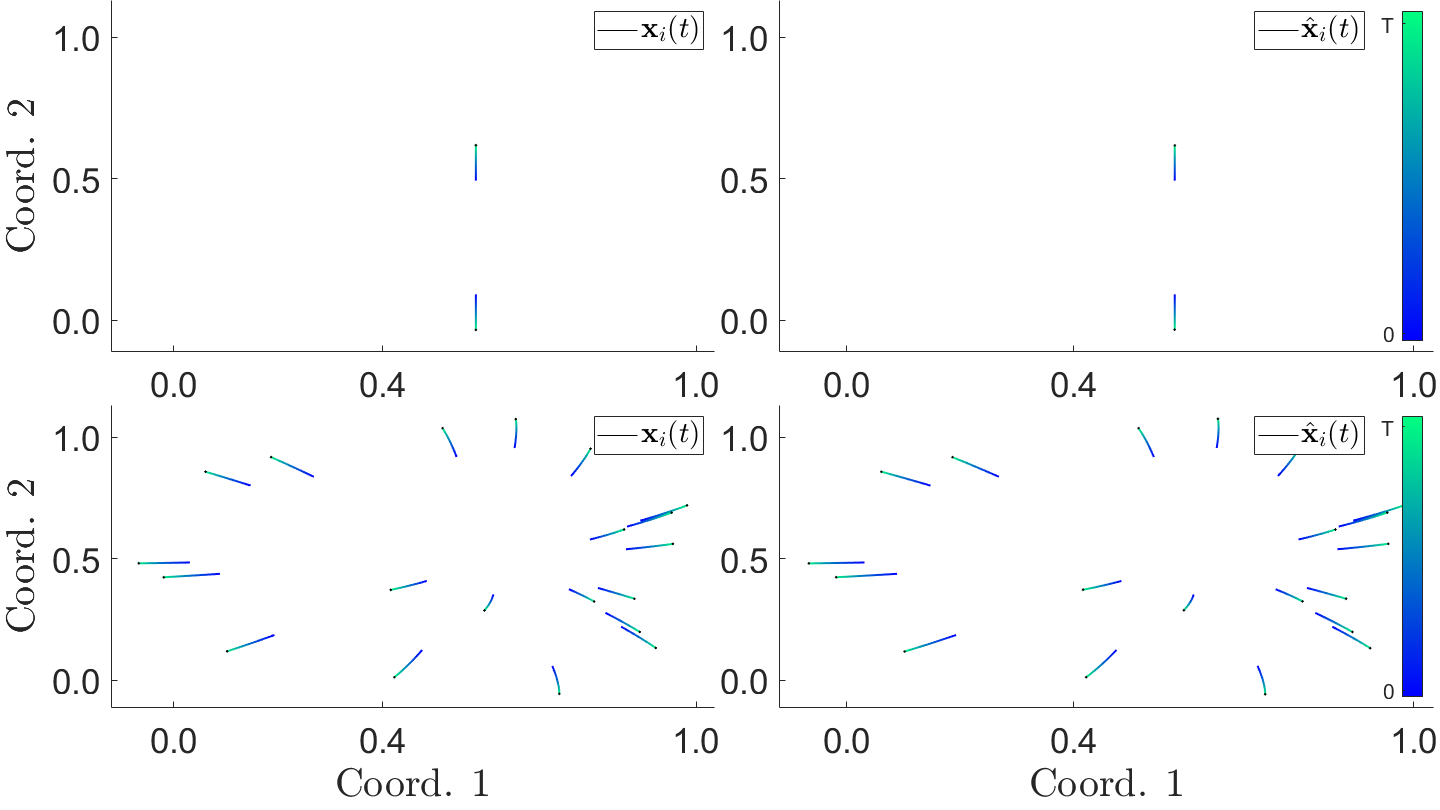}    
\caption{
{\em{Top}}: Comparison of the estimated and true interaction kernels (all of the plots are in the same $B\by$-coordinates for ease of comparison), demonstrating successful estimation even in the case of multiple variables, comparable to the Oracle case ($B$ known).
{\em{Bottom}}: $\{\bx_i(t)\}_{i = 1}^N$ vs. $\{\hat\bx_i(t)\}_{i = 1}^N$. Similar layout as in Fig. \ref{fig:OD_traj}.  We maintain a faithful reproduction of the dynamics, even for $N=20$, quantified in Table \ref{table:PLwDC_err}.} 
\label{fig:PLwDC_phi_traj_comp}
\end{center}
\end{figure}

\noindent{\bf{Power Law Dynamics with Directional Correction (PLwDC)}}.
We consider a direction-corrected power-law dynamics, generalizing that of \cite{KSUB2011}, where $\Phi(\bx_i, \bx_{i'}) = \phi(B\by(\bx_i, \bx_{i'}))$ with $B\by(\bx_i, \bx_{i'}) = ({\norm{\bx_i - \bx_{i'}}^2}/({2\sqrt{3}})$, ${\inp{\bx_{i'} - \bx_i, \bv_0}}/{\sqrt{2}})$ and $\phi(\xi_1, \xi_2) = (\sqrt{2\sqrt{3}\xi_1} - 1)\exp\left(\frac{2}{\pi}\arctan(\sqrt{2}\xi_2)\right)$. $\mu_0$ is as in the previous example.  Here $\bv_0 \in \R^d$ is a unit vector which defines a preferred direction for the agents to follow.  As shown in Table \ref{table:PLwDC_err}, our learning method can keep about $1$-digit relative accuracy in estimating the feature map and the interaction kernel -- a loss of accuracy compared to the $1D$ power-law case due to the increase in the dimension of the features.
The estimators of the interaction kernel are compared to the true kernel in Fig. \ref{fig:PLwDC_FM_dist_comp}.
We compare the distribution of $B\by$ and $\hat{B}\by$ in Fig. \ref{fig:PLwDC_FM_dist_comp}, $\phi$ to the two estimated pairs $(\hat{B}, \hat\phi_1)$ and $(B, \hat\phi_2)$ (Oracle estimator) as well as the trajectories in Fig. \ref{fig:PLwDC_phi_traj_comp}.

\begin{table}[H]
\begin{center}
\caption{PLwDC Errors}\label{table:PLwDC_err}
\scalebox{0.9}{
\begin{tabular}{l || l} 
$\text{Err}_{B}$ & $7.81 \cdot 10^{-1} \pm 1.4 \cdot 10^{-3}$ \\
\hline
$\text{Err}_{\phi}^{\text{rel}}$ & $1.22 \cdot 10^{-1} \pm 3.3 \cdot 10^{-4}$ \\
\hline
$\text{Err}_{\phi}^{\text{rel}}$ (Oracle) & $2.40 \cdot 10^{-1} \pm 3.6 \cdot 10^{-4}$ \\
\hline
$\text{Err}_{\text{traj}, \text{mean}}^{\text{rel}}$ & $2.8 \cdot 10^{-3} \pm 1.1 \cdot 10^{-4}$ \\
\hline
$\text{Err}_{\text{traj}, \text{mean}}^{\text{rel}}$ (Transfer) & $4 \cdot 10^{-3} \pm 1.4 \cdot 10^{-3}$\\
\end{tabular}
}
\end{center}
\vskip-0.25cm
\end{table}
Although our estimate $\hat B$ has large principal angle with $B$, we are able to provide an estimated pair $(\hat{B}, \hat\phi_1)$ with reasonably small $L^2(\rho_T)$ error (even smaller than knowing $B$), which also yields accurate trajectories. The reason lies in the unindentifiability of $B$, and the estimator produces variables which are in one-to-one correspondence with the ones in the model, see Fig. \ref{fig:PLwDC_FM_dist_comp}.
\section{Conclusion}\label{sec:conclude}
We have shown that we can estimate factorizable high dimensional interaction kernel $\Phi$ from high dimensional observation data by first learning the feature reduction map $B$ via MPLS, then inferring the reduced interaction kernel $\phi$ from a non-parametric regression with the reduction variable $\mbf{\xi}$.  Our method can accurately estimate the feature reduction map, i.e. the unknown reduced variables in the interaction kernel, as well as the reduced interaction kernel, and provide accurate predictions for the dynamics of the system.  Work on systems with different interaction structures, higher order systems, and transfer across multiple systems sharing reduced variables, is ongoing, as is the development of theoretical guarantees on the performance of the estimators, expecting to show that their consistency, under suitable identifiability assumptions, and convergence rate, not cursed by the dimension of the state space of the system nor by the rank of the feature map.

\noindent{{\bf{Acknolwedgments}}}.
MM thanks Fei Lu for insightful discussions.  
Prisma Analytics Inc. provided computing and storage facilities at no cost.
\bibliography{FML}                 
%
\appendix
\end{document}